\documentclass[journal]{IEEEtran}
\usepackage{bm}

\usepackage{amsfonts}
\usepackage{graphicx}
\usepackage{makecell}
\usepackage{tikz}
\usepackage{collcell}
\usepackage{colortbl}
\usepackage{pgfplots}
\usepackage{pgfplotstable}
\usepackage{multirow}
\usepackage{url}
\usepackage[all]{xy}
\usepackage{cite}
\usepackage{booktabs}
\usepackage{adjustbox}
\usepackage{tikz}
\usepackage{arydshln}
\usepackage{epstopdf}
\usepackage{amsmath}
\usepackage{float}
\usetikzlibrary{arrows,positioning}
\usetikzlibrary{plotmarks}
\usetikzlibrary{calc}

\usetikzlibrary{patterns}
\usetikzlibrary{shapes.geometric}
\pgfdeclarelayer{bg}    
\pgfsetlayers{bg,main}

\newcommand{\AlgorithmName}{3D-CAE}
\newcommand{\GaussianClustering}{GM}
\newcommand{\KMeans}{$k$-means}

\newcommand{\term}[1]{\mathfrak{#1}}
\newcommand{\zbior}[1]{\mathbb{#1}}  
\newcommand{\macierz}[1]{{\boldsymbol{\mathrm{#1}}}}
\newcommand{\wektor}[1]{\macierz{\MakeLowercase{#1}}}
\newcommand{\newItem}[2]{%
  \expandafter\def\csname #1\endcsname {\MakeLowercase{#2}} %
  \expandafter\def\csname n#1\endcsname {\MakeUppercase{#2}} %
  \expandafter\def\csname zb#1\endcsname {\zbior{\MakeUppercase{#2}}} %
  \expandafter\def\csname t#1\endcsname {\term{\MakeLowercase{#2}}} %
  \expandafter\def\csname m#1\endcsname {\macierz{\MakeUppercase{#2}}} %
  \expandafter\def\csname w#1\endcsname {\wektor{#2}} %
  }
  \newItem{Atrybut}{d_f}
\newItem{Deskryptor}{x}
\newItem{Regula}{l}
\newItem{Przeslanka}{a}
\newItem{Konsekwencja}{b}
\newItem{Implikacja}{b'}
\newItem{Zespolona}{c}
\newItem{Rzeczywista}{r}
\newItem{Klasa}{\delta}
\newItem{Decyzja}{y}
\newItem{Przyklad}{a}
\newItem{Test}{a}
\newItem{Klaster}{c}
\newItem{Rozmycie}{s}
\newItem{Centre}{v}
\newItem{Membership}{u}
\newItem{ClusterMembership}{\mu}
\newItem{Weight}{z}
\newItem{Krotka}{x}
\newItem{Width}{w}
\newItem{Wspolczynnik}{p}
\newItem{Obiekt}{k}
\newItem{Odleglosc}{t}
\newItem{Parameter}{p}

\ifCLASSINFOpdf
\else
\fi

\pgfplotstableset{
    /color cells/min/.initial=0,
    /color cells/max/.initial=1000,
    /color cells/textcolor/.initial=,
    color cells/.code={%
        \pgfqkeys{/color cells}{#1}%
        \pgfkeysalso{%
            postproc cell content/.code={%
                \begingroup
                \pgfkeysgetvalue{/pgfplots/table/@preprocessed cell content}\value
\ifx\value\empty
\endgroup
\else
                \pgfmathfloatparsenumber{\value}%
                \pgfmathfloattofixed{\pgfmathresult}%
                \let\value=\pgfmathresult
                \pgfplotscolormapaccess
                    [\pgfkeysvalueof{/color cells/min}:\pgfkeysvalueof{/color cells/max}]%
                    {\value}%
                    {\pgfkeysvalueof{/pgfplots/colormap name}}%
                \pgfkeysgetvalue{/pgfplots/table/@cell content}\typesetvalue
                \pgfkeysgetvalue{/color cells/textcolor}\textcolorvalue
                \toks0=\expandafter{\typesetvalue}%
                \xdef\temp{%
                    \noexpand\pgfkeysalso{%
                        @cell content={%
                            \noexpand\cellcolor[rgb]{\pgfmathresult}%
                            \noexpand\definecolor{mapped color}{rgb}{\pgfmathresult}%
                            \ifx\textcolorvalue\empty
                            \else
                                \noexpand\color{\textcolorvalue}%
                            \fi
                            \the\toks0 %
                        }%
                    }%
                }%
                \endgroup
                \temp
\fi
            }%
        }%
    }
}

\begin{document}
\title{Unsupervised Segmentation of Hyperspectral Images Using 3D Convolutional Autoencoders}
\author{Jakub Nalepa,~\IEEEmembership{Member,~IEEE}, 
        Michal Myller,
        Yasuteru Imai,
        Ken-ichi Honda,
        Tomomi Takeda,
        and Marek Antoniak
\thanks{This work was funded by European Space Agency in the HYPERNET project (JN, MM, MA). The research was carried out by HISUI (Hyperspectral Imager SUIte) public research promoted by the Ministry of Economy, Trade and Industry, Japan (YI, KH, TT).}
\thanks{J.~Nalepa and M.~Myller are with Silesian University of Technology, Gliwice, Poland (e-mail: jnalepa@ieee.org), JN, MM, and M.~Antoniak are with KP Labs, Gliwice, Poland (\{jnalepa, mmyller, mantoniak\}@kplabs.pl). Y. Imai and K. Honda are with Kokusai Kogyo, Co., Ltd., Tokyo, Japan (e-mail: \{yasuteru\_imai, kenichi\_honda\}@kk-grp.jp). T. Takeda is with Japan Space Systems, Tokyo, Japan (e-mail: takeda-tomomi@jspacesystems.or.jp).}
}


\maketitle
\begin{abstract}
Hyperspectral image analysis has become an important topic widely researched by the remote sensing community. Classification and segmentation of such imagery help understand the underlying materials within a scanned scene, since hyperspectral images convey a detailed information captured in a number of spectral bands. Although deep learning has established the state of the art in the field, it still remains challenging to train well-generalizing models due to the lack of ground-truth data. In this letter, we tackle this problem and propose an end-to-end approach to segment hyperspectral images in a fully unsupervised way. We introduce a new deep architecture which couples 3D convolutional autoencoders with clustering. Our multi-faceted experimental study---performed over benchmark and real-life data---revealed that our approach delivers high-quality segmentation without any prior class labels.
\end{abstract}

\begin{IEEEkeywords}
Hyperspectral imaging, unsupervised segmentation, deep learning, autoencoder, clustering.
\end{IEEEkeywords}

\IEEEpeerreviewmaketitle

\section{Introduction} \label{sec:intro}

Hyperspectral imaging (HSI) provides detailed information about the material within a captured scene. It registers a number of spectral bands, commonly up to hundreds of them, and can be exploited to understand the location and characteristics of the objects in the process of HSI \emph{classification} and \emph{segmentation}. In classification, we assign a single label to an input HSI pixel, whereas in segmentation we are focused on finding the boundaries of objects within an image\footnote{Therefore, segmentation involves classification of separate pixels.}. Due to the increased availability of hyperspectral sensors, HSI analysis has become an important research topic tackled by the machine learning, remote sensing, and pattern recognition communities. Such imagery has multiple applications in a plethora of fields, including biochemisty, biology, medicine, geosciences, military defense, food quality management and monitoring, pharmacy, and many more~\cite{8314827}. Hyperspectral imaging is an indispensable tool in Earth observation, as it captures Earth peculiarities that are useful in precision agriculture, managing environmental disasters, military defense applications, soil monitoring or prediction of environmental events.

HSI classification and segmentation techniques can be divided into conventional machine learning algorithms, requiring feature engineering (i.e.,~feature extraction and selection)~\cite{Bilgin2011,Dundar2018}, and deep learning-powered approaches, in which the appropriate representation is learned during the training~\cite{Chen2015,Zhao2016,Zhong2017,Mou2017,Santara2017,Lee2017,Gao_2018}. The success of deep learning is reflected in a variety of fields, where it established the state of the art. However, to deploy deep models in practice, we need large and representative ground-truth training sets. It is a serious limiting factor in hyperspectral Earth observation analysis, where transferring HSI from an imaging satellite back to Earth is extremely costly. Creating new ground-truth datasets is error-prone and requires building a thorough understanding of the materials within a scene. Therefore, it involves acquiring observational ground-sensor data---it is often cost- and time-inefficient. These difficulties result in a very small number of ground-truth HSIs. In~\cite{Nalepa2019GRSL}, we analyzed 17 recent papers in which only \emph{seven} benchmarks were exploited, with only \emph{three} of them being ``widely-used'': Pavia University (15 papers), Indian Pines (in 8 papers), and Salinas Valley (5 papers).

There are three main approaches to deal with the limited ground-truth hyperspectral sets: ({i})~\emph{data augmentation}, ({ii})~\emph{transfer learning}, and ({iii})~\emph{unsupervised} analysis of HSI, with (i) and (ii)~being exploited mostly in deep learning-powered techniques. Data augmentation is a process of generating artificial examples following the original data distribution. Such samples can extend the training sets or they can be elaborated at the inference time, to build an intrinsic ensemble-like deep model~\cite{8746168}. In transfer learning, we train feature extractors over the \emph{source} training data, and apply it to the \emph{target} data~\cite{DBLP:journals/corr/abs-1906-09631}. This approach allows us to benefit from the available data to train efficient extractors---the classification part of a network is later fine-tuned over the target (much smaller) HSI. Both augmentation and transfer learning require the annotated target sets to either input them to an augmentation engine, or to utilize them for fine-tuning the deep models. Hence, their usefulness is limited in scenarios where manually-analyzed HSIs do not exist and are infeasible to generate.

\begin{figure*}[ht!]
\centering
\begin{tabular}{c}
  \hspace*{0.0cm}\includegraphics[width=0.805\paperwidth]{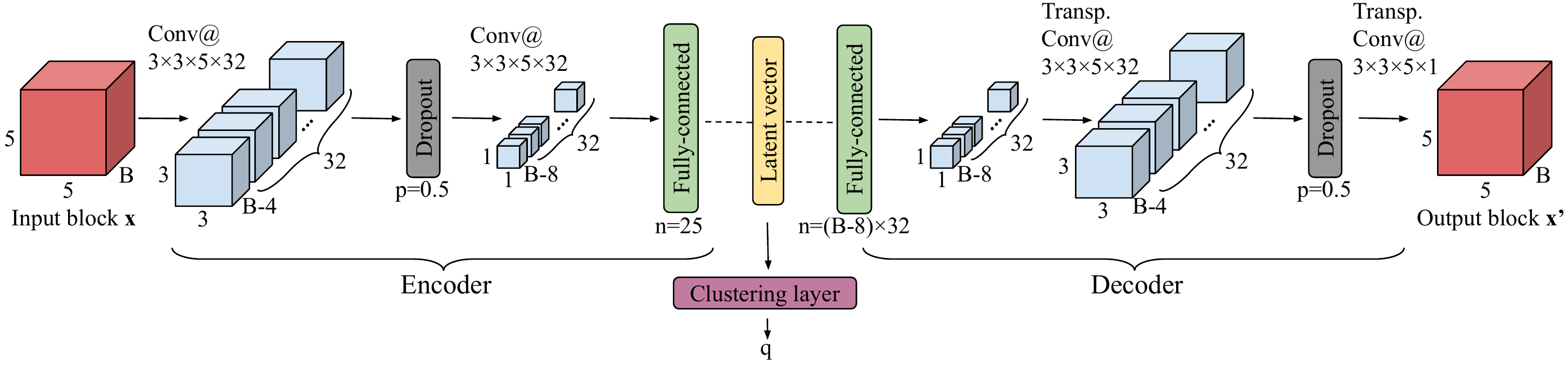}\\
\end{tabular}
	\caption{Our 3D convolutional autoencoder coupled with a clustering layer is trained in two stages---first, we learn a latent data representation (the clustering layer is not used in this stage, and the loss reflects the data reconstruction abilities of the 3D convolutional autoencoder), and then we focus on clustering while still allowing for improvements in the latent representations by incorporating the clustering loss into the loss function.}
	\label{fig:architecture}
\end{figure*}

On the other hand, unsupervised segmentation offers the possibility of processing HSI \emph{without} any prior class labels. Although the literature in unsupervised HSI segmentation is rather limited, there exist conventional machine learning approaches (exploiting hand-crafted features) which benefit from mean shift filtering~\cite{SPIE_Unsupervised}, diffusion-based dimensionality reduction followed by clustering~\cite{Schclar2019}, and phase-correlation analysis\cite{1715310}. In~\cite{8082108}, the authors used a fully conv-deconv deep network for unsupervised spectral-spatial feature learning. Then, the convolutional subnetwork was used as a generic feature extractor over the target (labeled) data in the transfer learning approach. A similar technique of extracting deep features using stacked sparse autoencoders, and later embedding them into linear support vector machines has been proposed in~\cite{7296592}.

In this letter, we tackle the problem of limited ground-truth hyperspectral sets, and propose a deep learning technique for unsupervised HSI segmentation. Inspired by a recent work by Guo et al.~\cite{10.1007/978-3-319-70096-0_39}, we introduce a 3D convolutional autoencoder architecture to learn embedded features, which later undergo clustering (Section~\ref{sec:method}). This clustering is performed \emph{during} the network training with a clustering-oriented loss, therefore our method delivers end-to-end unsupervised HSI segmentation. To the best of our knowledge, such approaches have not been investigated in the HSI literature so far. We performed a multi-faceted experimental study---over benchmark and real-life hyperspectral data---to understand the abilities of the proposed technique. It showed that our method offers high-quality and consistent segmentation, and does not require any prior class labels to effectively segment HSI (Section~\ref{sec:experiments}).

\section{Unsupervised HSI Segmentation Using 3D Convolutional Autoencoders}\label{sec:method}

In our unsupervised HSI segmentation approach (3D-CAE) inspired\footnote{In~\cite{10.1007/978-3-319-70096-0_39}, the authors used much deeper architectures without dropout over full input images (with 2D kernels only) in the context of image classification.} by~\cite{10.1007/978-3-319-70096-0_39}, we exploit 3D convolutional autoencoders (CAEe) to extract deep features which later undergo clustering (Fig.~\ref{fig:architecture}). In the encoding part of the network, we capture both spectral and spatial features within an input three-dimensional hyperspectral patch $\bm{x}$ (of size $5\times 5\times {\rm B}$, where ${\rm B}$ is the number of bands; the patches are extracted with unit stride) using two convolutional layers denoted as ${\rm Conv}@h_k\times w_k\times d_k\times k$, for which we define the height ($h_k$), width ($w_k$), and depth ($d_k$) of the kernels, alongside the number of kernels in this layer (here, $k=32$ for all convolution/transposed convolution layers)---each kernel moves with unit stride in each direction. These convolutional layers are interleaved with one dropout layer (with the dropout probability of $p=0.5$) acting as a regularizer. The central-pixel features in the patch are later re-shaped to form a 1D vector which becomes an input to a fully-connected (embedding) layer with $n=25$ neurons, whose output is the latent vector. These embedded features are transformed back to the original 3D patch (to get the output 3D patch $\bm{x'}$) in the decoding part of a CAE which is a mirrored version of the encoder with the transposed convolutions applied for up-sampling. The CAE is learned in the first training stage with the following reconstruction loss:
\begin{equation}
L_r=\frac{1}{p}\sum_{i=1}^{p}\left|\left|\bm{x}_i-\bm{x'}_i\right|\right|^2_2,
\end{equation} where $p$ is the number of 3D \emph{patches} in a batch. This stage, in which we do \emph{not} use the clustering layer, runs until reaching convergence or the stopping condition (in this letter, the optimization terminates if the difference between two consecutive reconstruction loss values is less than $\epsilon=10^{-6}$).

In the second training stage, we modify the loss function and ``switch on'' the clustering layer---it is connected to the embedded layer of CAE which outputs the latent vector $z_i$ for the $i$-\textit{th} input patch. The embedded features are being assigned a soft label $q_i$ in the clustering layer. As proposed in~\cite{10.1007/978-3-319-70096-0_39}, this layer maintains the cluster centers $\mu_j$, where $j=1,2,\dots J$, and $J$ is the number of clusters, as trainable weights. The probability of assigning an input 3D patch $\bm{x}_i$ to each $j$-\textit{th} cluster is generated using the Student's $t$-distribution:
\begin{equation}
  q_{ij}=\frac{(1+\left|\left|z_i-\mu_j\right|\right|^2)^{-1}}{\sum_{j=1}^{J}\left(1+\left|\left|z_i-\mu_j\right|\right|^2\right)}.
\end{equation}
\noindent Finally, the clustering loss is:
\begin{equation}
  L_c={\rm KL}(\mathcal{T}||q)=\sum_{i=1}^{p'}\sum_{j=1}^{J}t_{ij}\log \frac{t_{ij}}{q_{ij}},
\end{equation}
\noindent where $p'$ is the number of \emph{pixels} in the batches, KL is the Kullback-Leibler divergence, and $\mathcal{T}$ is the target distribution:
\begin{equation}
 t_{ij}=\frac{q_{ij}^2/\sum_{i=1}^{p'}q_{ij}}{\sum_{j=1}^{J}\left(q_{ij}^2/\sum_{i=1}^{p'}q_{ij}\right)}.
\end{equation}
\noindent The clustering loss is incorporated into the total loss function $L$ used in this training stage, and it becomes:
\begin{equation}
L=L_r+\alpha L_c,
\end{equation}
\noindent where $0<\alpha<1$, and it is a loss weighting coefficient (we used $\alpha=0.1$). This stage continues until the convergence or the termination condition is met (we restrict it to 25 epochs).

\section{Experiments}\label{sec:experiments}

\begin{figure}[ht!]

    \newcommand{\dd}{0.8}
\newcommand{\ddx}{0.02}
\newcommand{\ws}{0.085}
\newcommand{\shiftx}{0.1}
\newcommand{\shifty}{0}
\definecolor{mycolor}{RGB}{176,213,169}
        \centering
        \resizebox{0.8\columnwidth}{!}{%
            \begin{tikzpicture}[scale=0.8,
                    input/.style={rectangle,rounded corners,draw=black,fill=red!20,inner sep=5pt,minimum height=50pt,minimum width=10pt,text width=5pt,text badly centered,thick},
                    output/.style={rectangle,rounded corners,draw=black,fill=white,inner sep=5pt,minimum height=50pt,minimum width=10pt,text width=5pt,text badly centered,thick},
                    convolution/.style={rectangle,rounded corners,draw=black,fill=cyan!20,inner sep=5pt,minimum height=50pt,minimum width=10pt,text width=5pt,text badly centered,thick},
                    maxpooling/.style={rectangle,rounded corners,draw=black,fill=orange!20,inner sep=5pt,minimum height=50pt,minimum width=10pt,text width=5pt,text badly centered,thick},
                    batchNorm/.style={rectangle,rounded corners,draw=black,fill=blue!20,inner sep=5pt,minimum height=50pt,minimum width=10pt,text width=5pt,text badly centered,thick},
                    fullyconnected/.style={rectangle,rounded corners,draw=black,fill={mycolor},inner sep=5pt,minimum height=50pt,minimum width=10pt,text width=5pt,text badly centered,thick},
                    softmax/.style={rectangle,rounded corners,draw=black,fill={mycolor},inner sep=5pt,minimum height=50pt,minimum width=10pt,text width=5pt,text badly centered,thick},
                    myarrow/.style={thick},
                    dottedarrow/.style={dotted, thick}]

                \newcommand{\HSIpixel}{{\rotatebox{90}{{HSI pixel}}}}
                \newcommand{\class}{{\rotatebox{90}{{\textbf{Label}}}}}
                \newcommand{\sep}{20}
                \newcommand{\bigSep}{20}
                \newcommand{\convolution}{{\rotatebox{90}{{Conv}}}}
                \newcommand{\maxpooling}{{\rotatebox{90}{{Max pool}}}}
                \newcommand{\batchnormalization}{{\rotatebox{90}{{BN}}}}
                \newcommand{\deconvolution}{{\rotatebox{90}{{Deconv}}}}
                \newcommand{\fullyconnected}{{\rotatebox{90}{{FC}}}}
                \newcommand{\softmax}{{\rotatebox{90}{{Softmax}}}}
                \newcommand{\convDescrSep}{1}
                \mathchardef\mhyphen="2D 

                \node (in) [input] {$\HSIpixel$};
                \node (conv1) [convolution, right=30 pt of in]{\convolution};
                \node (convDescription) at ($(conv1.south) - (0 pt, \convDescrSep pt)$)[anchor=north] {{\tiny$\bm{s=1\times 1\times 5}$}};
                \node (convDescription2) at ($(convDescription.south) + (0 pt, 5 pt)$)[anchor=north] {{\tiny$\bm{n=200}$}};
                \node (batchnorm1) [batchNorm, right=\sep pt of conv1]{\batchnormalization};
                \node (maxpooling1) [maxpooling, right=\sep pt of batchnorm1]{\maxpooling};
                \node (convDescription3) at ($(maxpooling1.south) - (0 pt, \convDescrSep pt)$)[anchor=north] {{\tiny$\bm{s=2\times 2}$}};
                \node (fc1) [fullyconnected, right=\sep pt of maxpooling1]{\fullyconnected};
                \node (convDescription3) at ($(fc1.south) - (0 pt, \convDescrSep pt)$)[anchor=north] {{\tiny$\bm{l_1=512}$}};
                \node (convDescription4) at ($(convDescription3.south) + (0 pt, 5 pt)$)[anchor=north] {{\tiny$\bm{l_2=128}$}};
                \node (softmax1) [softmax, right=\sep pt of fc1]{\softmax};
                \node (out) [output, right=\sep pt of softmax1]{\class};


                \draw[->,myarrow] (in) -- (conv1);
                \draw[->,myarrow] (conv1) -- (batchnorm1);
                \draw[->,myarrow] (batchnorm1) -- (maxpooling1);
                \draw[->,myarrow] (maxpooling1) -- (fc1);
                \draw[->,myarrow] (fc1) -- (softmax1);
                \draw[->,myarrow] (softmax1) -- (out);
            \end{tikzpicture}
        } \vspace*{-0.2cm}\caption{1D-CNN with $n$ kernels in the convolutional layer ($s$ stride) and $l_1$ and $l_2$ neurons in the fully-connected (FC) layers. BN is batch normalization.} \label{fig:1d_network}
    \end{figure}
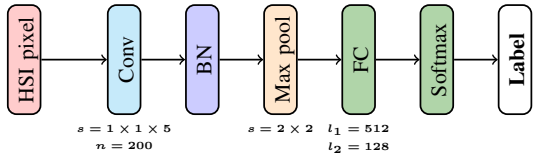

The objectives of our experiments are multi-fold. We verify the abilities of our unsupervised classification technique and compare it with other state-of-the-art clustering methods: $k$-means, where $k$ equals the number of target classes in the benchmark sets, and Gaussian mixture modeling (being a generalization of $k$-means which incorporates information about the covariance structure of the data and the centers of latent Gaussians), applied over the original (full) and reduced HSI. In the latter case, we reduce the dimensionality of the input HSI to match the size of our latent vector using (i)~principal component analysis (PCA), (ii)~independent component analysis (ICA), (iii)~our sliding-window algorithm for simulating multispectral imagery from its hyperspectral counterpart, in which we generate the averaged band within a non-overlapping sliding window (S-MSI)~\cite{Marcinkiewicz2019IGARSS}, and (iv)~our CAE. In this letter, the latent vector is of size $25$ (Fig~\ref{fig:architecture}). Also, we apply our CAE over reduced HSI and check the impact of the dimensionality reduction on its abilities (in this case, CAEs \emph{do not} perform dimensionality reduction, and the latent vector is of the same size as the input vector). Finally, we compare unsupervised segmentation with our 1D-CNN~\cite{Nalepa2019GRSL} (Fig.~\ref{fig:1d_network}) trained in a supervised manner over original and reduced benchmarks. Our study was divided into two experiments, over the available benchmarks (Section~\ref{sec:experiment1}), and a real-life HSI for which the ground-truth segmentation does not exist (Section~\ref{sec:experiment2}).

\begin{figure}[ht!]
\centering
\hspace*{-0.1cm}
\setlength\tabcolsep{1pt}
\begin{tabular}{cccc}
  \includegraphics[width=0.89\columnwidth]{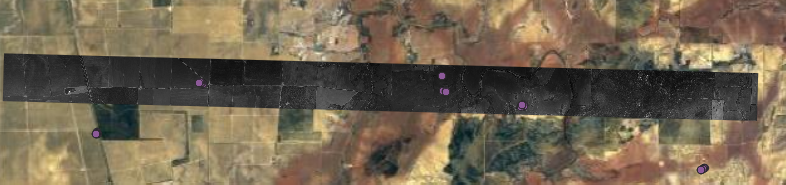}\\
\end{tabular}
	\caption{The Mullewa area with an annotated region which is captured by the HSI segmented in this letter. The violet points indicate the test fields where the wheat in-situ measurements had been performed.}
	\label{fig:mullewa_vis}
\end{figure}

We exploited three most popular HSI benchmarks from the literature (\url{http://www.ehu.eus/ccwintco/index.php/Hyperspectral_Remote_Sensing_Scenes}): (i)~Salinas Valley (Sa), USA ($217\times 512$ pixels, AVIRIS sensor) showing different sorts of vegetation (16 classes, 224 bands, 3.7 m spatial resolution); (ii)~Indian Pines (IP), USA ($145\times 145$, AVIRIS)---agriculture and forest (16 classes, 200 bands, 20~m); (iii)~Pavia University (PU), Italy ($340\times 610$, ROSIS)---urban scenery (9 classes, 103 channels, 1.3 m). We also utilize the aerial hyperspectral observations acquired using the HyMap airborne sensor ($7982\times 512$, HyVista Corp. Pty Ltd., Australia, 126 bands with a wavelength resolution of 20 nm, 4.2 m) on Oct. 29, 2009. The study area was located in Mullewa, Western Australia (480 km$^2$), and it is mainly used for the wheat, canola and lupin production. Although there were 30 test fields in which in-situ measurements had been performed (captured 1 m above a head of the wheat, eight measurements at each point; the measurement points are rendered in violet in Fig.~\ref{fig:mullewa_vis}), such data is not suitable for verifying segmentation algorithms, as we know the class label of an extremely small subset of all pixels. Hence, for Mullewa, we focus on qualitative analysis.

We use two clustering-quality measures to quantify the performance of the unsupervised techniques: normalized mutual information (NMI) and adjusted rand score (ARS). NMI is:
\begin{equation}
  {\rm NMI}=\frac{{\rm MI}(A,B)}{\left[H(A)+H(B)\right]/2},
\end{equation}
\noindent where ${\rm MI}(A, B)=H(A)+H(B)-H(A,B)$ is the mutual information index quantifying the value of information shared between two random variables $A$ and $B$, $H(\cdot)$ denotes entropy, and $H(A,B)$ is the joint entropy between clusterings. ARS is:
\begin{equation}
  {\rm ARS}=\frac{\binom{n}{2}\left(a+d\right)-\left[\left(a+b\right)\cdot\left(a+c\right)+\left(c+d\right)\cdot\left(b+d\right)\right]}{\binom{n}{2}^2-\left[\left(a+b\right)\cdot\left(a+c\right)+\left(c+d\right)\cdot\left(b+d\right)\right]},
\end{equation}
\noindent where $n$ is the number of objects (pixels) subjected to clustering, and $a$, $b$, $c$, and $d$ denote the number of data points placed in: the same group (cluster) in $A$ and $B$ ($a$), the same groups in $A$ and in different groups in $B$ ($b$), the same groups in $B$ and in different groups in $A$ ($c$), and in different groups in $A$ and in different groups in $B$ ($d$)~\cite{10.1007/978-3-642-04277-5_18}. Both NMI and ARS range from 0 to 1, where 1 means perfect score. Additionally, for 1D-CNN trained in a supervised manner using our balanced division into the training and validation sets (as presented in~\cite{Nalepa2019GRSL}; the B division), we report the average accuracy (AA), overall accuracy (OA), and the kappa scores $\kappa=1-\frac{1-p_o}{1-p_e}$, where $p_o$ and $p_e$ are the relative observed agreement, and hypothetical probability of chance agreement, respectively, and $-1\leq \kappa\leq 1$ ($\kappa=1$ is the perfect score). These scores were obtained over the \emph{entire} input HSI to make them comparable with the unsupervised segmentation performed over the entire HSI (in both cases, however, we exclude the background pixels for which the class label is unknown), and averaged across 30 runs. Since the test set for 1D-CNN includes the training and validation examples, the results can be considered over-optimistic~\cite{Nalepa2019GRSL}. The deep networks were coded in \texttt{Python 3.6}, and the supervised training of 1D-CNN (ADAM, learning rate of $10^{-4}$, $\beta_1 = 0.9$, $\beta_2 = 0.999$) terminated, if after 25 epochs OA over the validation set (random subset of the training set) does not change. The experiments ran on NVIDIA GeForce RTX~2080.

\subsection{Experiment 1: Benchmark data}\label{sec:experiment1}

In this experiment, we compare 3D-CAE with other techniques over three HSI benchmarks---each unsupervised approach was executed exactly \emph{once}, in order to understand its real-life applicability, where running algorithms multiple times is infeasible. Also, we performed Monte-Carlo cross-validation (repeated $30\times$) with balanced training and validation sets~\cite{Nalepa2019GRSL}, and analyze the average supervised measures (AA, OA, and $\kappa$) obtained using 1D-CNN. For the sake of completeness, we report the unsupervised segmentation measures (NMI and ARS) for 1D-CNN as well---the \emph{entire} scene (without background) was segmented. Since the test set includes both training and validation sets in this case (there is an ``training-test information leak''), NMI and ARS may be considered over-optimistic for 1D-CNN.

\begin{table}[ht!]
	\scriptsize
	\centering
	\caption{The results obtained over all benchmarks.}
	\label{tab:clustering_measure}
	\renewcommand{\tabcolsep}{0.25cm}
\newcommand{\mybackground}{35}
	\begin{tabular}{l|rr|rr|rr}
\multicolumn{7}{c}{\textbf{Unsupervised segmentation measures}}\\
 \hline
\multicolumn{1}{r|}{Set$\rightarrow$} & \multicolumn{2}{c|}{Sa} & \multicolumn{2}{c|}{IP} & \multicolumn{2}{c}{PU} \\
\hline
Algorithm$\downarrow$ & NMI & ARS & NMI & ARS & NMI & ARS \\
\hline

1D-CNN*                            &0.885                       &0.725                      &0.705                      &0.586                      &\cellcolor{green!40}0.786  &\cellcolor{green!40}0.771\\
1D-CNN*~with PCA                   &\cellcolor{gray!\mybackground}0.860                       &\cellcolor{gray!\mybackground}0.685                      &\cellcolor{gray!\mybackground}0.640                      &\cellcolor{gray!\mybackground}0.493                      &\cellcolor{gray!\mybackground}0.360  &\cellcolor{gray!\mybackground}0.215\\
1D-CNN*~with ICA                   &0.873                       &0.723                      &0.641                      &0.502                      &0.616  &0.556\\
1D-CNN*~(S-MSI)                    &0.880                       &0.723                      &\cellcolor{green!40}0.718  &\cellcolor{green!40}0.609  &0.784  &0.759\\
1D-CNN*~(CAE)                      &\cellcolor{green!40}0.886   &\cellcolor{green!40}0.738  &0.663                      &0.496                      &0.748  &0.658\\
\hline
\GaussianClustering               &0.819&0.642&0.445&0.229&0.514&\cellcolor{gray!\mybackground}0.290\\
\GaussianClustering~with PCA      &0.830&0.654&0.443&0.235&0.530&0.404\\
\GaussianClustering~with ICA      &0.838&0.665&0.436&\cellcolor{gray!\mybackground}0.212&0.522&0.396\\
\GaussianClustering~(S-MSI)       &\cellcolor{green!40}\textbf{0.848}&\cellcolor{green!40}\textbf{0.673}&\cellcolor{green!40}0.456&0.248&\cellcolor{green!40}0.532&0.407\\
\GaussianClustering~(CAE)         &\cellcolor{red!30}0.628&\cellcolor{red!30}0.475&\cellcolor{gray!\mybackground}0.435&\cellcolor{green!40}0.289&\cellcolor{gray!\mybackground}0.480&\cellcolor{green!40}0.459\\
\hdashline
\KMeans                           &\cellcolor{green!40}0.732&\cellcolor{green!40}0.538&0.437&0.211&\cellcolor{green!40}0.546&\cellcolor{green!40}0.350\\
\KMeans~with PCA                  &0.724&0.524&0.430&0.204&0.545&0.324\\
\KMeans~with ICA                  &0.730&0.535&\cellcolor{red!30}0.381&\cellcolor{red!30}0.178&\cellcolor{red!30}0.477&\cellcolor{red!30}0.263\\
\KMeans~(S-MSI)                   &0.712&\cellcolor{gray!\mybackground}0.496&0.430&0.208&\cellcolor{green!40}0.546&0.325\\
\KMeans~(CAE)                     &\cellcolor{gray!\mybackground}0.710&0.503&\cellcolor{green!40}0.451&\cellcolor{green!40}\textbf{0.297}&0.539&0.336\\
\hdashline
\textbf{\AlgorithmName}           &\cellcolor{gray!\mybackground}0.714&0.533&\cellcolor{gray!\mybackground}0.431&\cellcolor{gray!\mybackground}0.231&0.553&0.339\\
\textbf{\AlgorithmName}~with PCA  &0.746&\cellcolor{gray!\mybackground}0.527&0.467&0.263&\cellcolor{green!40}\textbf{0.639}&\cellcolor{green!40}\textbf{0.546}\\
\textbf{\AlgorithmName}~with ICA  &\cellcolor{green!40}0.839&\cellcolor{green!40}0.644&\cellcolor{green!40}\textbf{0.504}&\cellcolor{green!40}0.278&\cellcolor{gray!\mybackground}0.538&\cellcolor{gray!\mybackground}0.316\\
\textbf{\AlgorithmName}~(S-MSI)   &0.728&0.531&0.442&0.241&0.601&0.450\\

\hline
	\end{tabular}

\scriptsize
	\renewcommand{\tabcolsep}{0.09cm}
	\begin{tabular}{l|rrr|rrr|rrr}
\multicolumn{10}{c}{\textbf{Supervised segmentation measures}}\\
 \hline
\multicolumn{1}{r|}{Set$\rightarrow$} & \multicolumn{3}{c|}{Sa} & \multicolumn{3}{c|}{IP} & \multicolumn{3}{c}{PU} \\
\hline
Algorithm$\downarrow$ & AA & OA & $\kappa$ & AA & OA & $\kappa$ & AA & OA & $\kappa$ \\
\hline

1D-CNN            &0.946&0.887&0.875&0.828&0.777&0.749&\cellcolor{green!40}0.894&\cellcolor{green!40}0.872&\cellcolor{green!40}0.835\\
1D-CNN~with PCA   &\cellcolor{gray!\mybackground}0.873&\cellcolor{gray!\mybackground}0.820&\cellcolor{gray!\mybackground}0.802&\cellcolor{gray!\mybackground}0.766&\cellcolor{gray!\mybackground}0.691&0.655&\cellcolor{gray!\mybackground}0.451&\cellcolor{gray!\mybackground}0.398&\cellcolor{gray!\mybackground}0.326\\
1D-CNN~with ICA   &\cellcolor{green!40}0.953&\cellcolor{green!40}0.904&\cellcolor{green!40}0.893&0.803&0.736&0.702&0.771&0.713&0.645\\
1D-CNN~(S-MSI)    &0.943&0.887&0.874&\cellcolor{green!40}0.832&\cellcolor{green!40}0.790&\cellcolor{green!40}0.762&0.875&0.839&0.796\\
1D-CNN~(CAE)      &0.946&0.895&0.875&0.812&0.735&\cellcolor{gray!\mybackground}0.650&0.876&0.822&0.764\\

\hline
\multicolumn{10}{l}{\scriptsize \textbf{How to read this table:} The \emph{globally} best \emph{unsupervised} method is boldfaced.}\\
\multicolumn{10}{l}{\scriptsize The background of the \emph{globally} worst \emph{unsupervised} method is red.}\\
\multicolumn{10}{l}{\scriptsize For each method, we annotate its best and worst variant (green and gray background).}\\
\multicolumn{10}{l}{\scriptsize *For the sake of completeness, we report the \emph{unsupervised} measures obtained using}\\
\multicolumn{10}{l}{\scriptsize 1D-CNN trained in a \emph{supervised} setting.}\\
	\end{tabular}

\end{table}

In Table~\ref{tab:clustering_measure}, we gather the experimental results obtained over all sets. They show that 3D-CAE consistently delivers high-quality segmentation in all settings, with and without HSI reduction (in all cases, we decrease the feature dimensionality to $25$ to match the number of 3D-CAE embedded features). On the other hand, the dimensionality reduction is beneficial in the unsupervised setting, and leads to better clustering. It indicates that only a small portion of the entire spectrum conveys useful information about the captured materials within those HSI---exploiting the full spectrum makes segmentation much harder due to the curse of dimensionality (the best results were obtained using our S-MSI; Wilcoxon test, $p<0.001$). These observations are confirmed in Table~\ref{tab:ranking}, where we report the ranking of all methods averaged across the benchmarks.

The execution time of all unsupervised techniques is reported in Table~\ref{tab:ranking}. These times reflect \emph{only} segmentation, without feature extraction for the methods run over reduced HSI (PCA took 1.17~s, 0.35~s, and 1.95~s for Sa, IP, and PU, respectively, ICA: 25.04~s, 1.28~s, and 56.68~s, S-MSI: 0.28~s, 0.04~s, 0.47~s, and 3D-CAE: 2843.27~s, 273.51~s, and 2975.91~s). Although 3D-CAE was significantly slower than other algorithms\footnote{The execution time of 3D-CAE over reduced HSI was very consistent with other deep learning-powered HSI segmentation techniques~\cite{8082108}.}, it retrieved consistently better segmentation (Table~\ref{tab:clustering_measure}). Also, we did not exploit early stopping for the clustering phase of 3D-CAE (it ran always for 25 epochs). This part of the training could have been terminated much earlier (as the training converged), which could have greatly reduced its execution time. It however requires further investigation.

\begin{table}[ht!]
	\scriptsize
	\centering
	\caption{The ranking (averaged across all benchmarks), and the execution time of all methods. The best ranking is boldfaced.}
	\label{tab:ranking}
	\renewcommand{\tabcolsep}{0.21cm}
	\begin{tabular}{l|rr|rrrrrrr}
 \hline
 & \multicolumn{2}{c|}{Ranking} & \multicolumn{4}{c}{Time (min)}\\
 \hline
Algorithm$\downarrow$ & NMI & ARS & Sa & IP & PU & Mu \\
\hline
\GaussianClustering & 7.33	&9.00 & 11.83	&1.63	&2.89	&91.49	\\
\GaussianClustering~with PCA & 6.67	&5.00 & 1.35	&0.09	&1.65&	32.97\\
\GaussianClustering~with ICA &7.67	&6.00	& 0.87	&0.29	&0.50	&5.40\\
\GaussianClustering~(S-MSI)&4.33	&\textbf{3.33}	& 1.03&	0.15&	1.41&	21.33 \\
\GaussianClustering~(CAE)&12.33	&6.00	& 0.98 & 0.18 & 0.92 & 18.42 \\
\hdashline
\KMeans&6.50	&8.00 & 1.12&	0.18&	0.79&	52.01\\
\KMeans~with PCA&9.50	&11.67 & 0.28	&0.07&	0.37	&16.96\\
\KMeans~with ICA&12.00	&11.67	& 0.35	&0.04	&0.99&	13.84\\
\KMeans~(S-MSI)&9.67	&11.67	& 0.31	&0.06	&0.34&	19.75\\
\KMeans~(CAE)&8.00	&7.33 & 0.26 & 	0.08 & 0.44 & 19.82\\
\hdashline
\textbf{\AlgorithmName}&8.33	&8.00	& 91.31&	14.75&	102.12	&1994.20\\
\textbf{\AlgorithmName}~with PCA&\textbf{3.00}	&5.00	& 20.15&	7.18&	36.28&	640.45\\
\textbf{\AlgorithmName}~with ICA&3.67	&6.33	& 16.88&	5.37&	27.99&	573.52\\
\textbf{\AlgorithmName}~(S-MSI)&6.00	&6.00	& 20.33	&5.36	&50.60	&553.55 \\
\hline
	\end{tabular}
\end{table}

\subsection{Experiment 2: Real-life data}\label{sec:experiment2}

\begin{figure*}[ht!]
\centering
\newcommand{\mywidth}{1.95}
\hspace*{-0.4cm}
\setlength\tabcolsep{1pt}
\begin{tabular}{ccc}
  &{\rotatebox{90}{\scriptsize GM}}& \includegraphics[width=\mywidth\columnwidth]{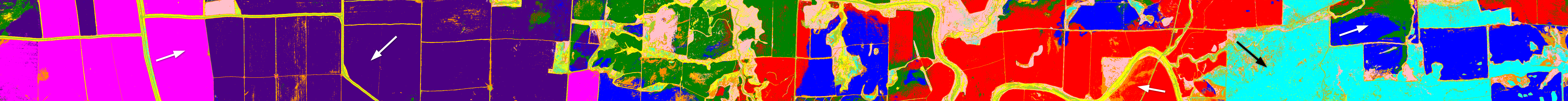}\\
  {\rotatebox{90}{\centering \scriptsize GM }}&{\rotatebox{90}{\scriptsize \centering(S-MSI)}}&\includegraphics[width=\mywidth\columnwidth]{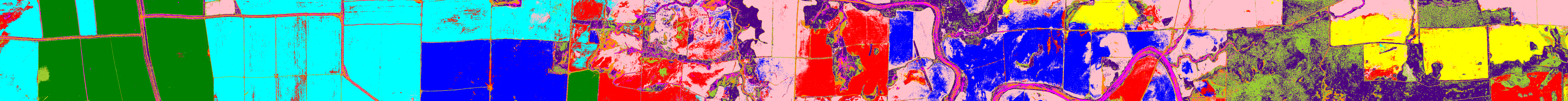}\\
  & {\rotatebox{90}{\scriptsize $k$-means}}&\includegraphics[width=\mywidth\columnwidth]{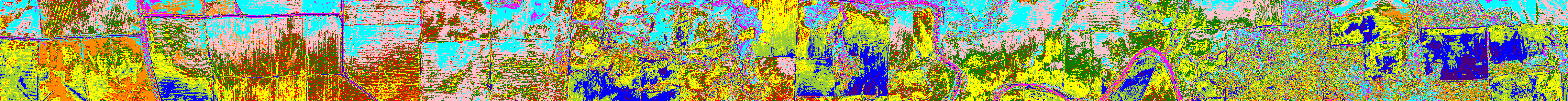}\\
  {\rotatebox{90}{\scriptsize $k$-means}}&{\rotatebox{90}{\scriptsize (CAE)}}&\includegraphics[width=\mywidth\columnwidth]{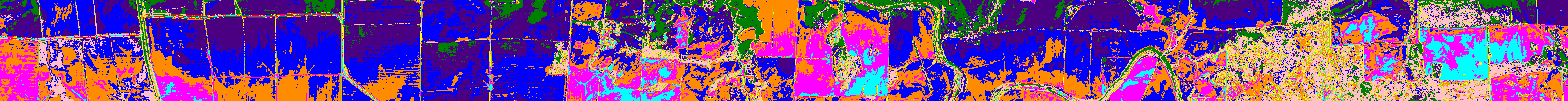}\\
  & {\rotatebox{90}{\scriptsize \textbf{3D-CAE}}}&\includegraphics[width=\mywidth\columnwidth]{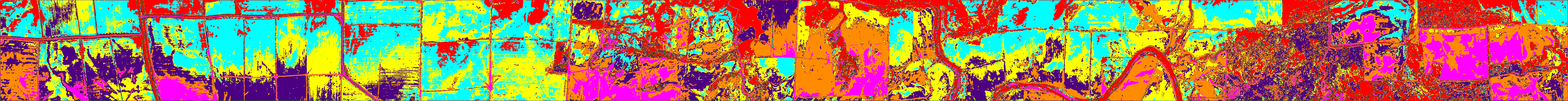}\\
  {\rotatebox{90}{\scriptsize \textbf{3D-CAE}}}& {\rotatebox{90}{\scriptsize \textbf{with PCA}}}&\includegraphics[width=\mywidth\columnwidth]{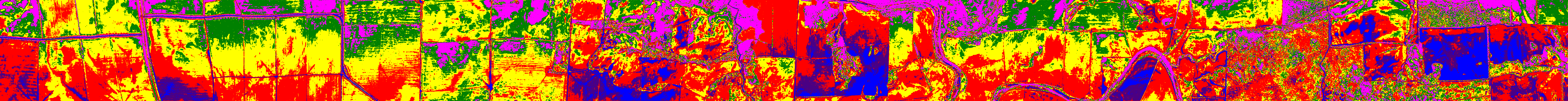}\\
\end{tabular}
	\caption{Segmentation of Mullewa obtained using the selected variants of all investigated techniques (for all visualizations in high-resolution, see \protect\url{https://gitlab.com/jnalepa/3d-cae}). The white and black arrows show the areas which are ``heterogeneous'' in $k$-means and 3D-CAE. It is in contrast to GM---it may indicate that GM had not appropriately captured subtle spectral differences within those regions (which finally were annotated as single-class regions).}
	\label{fig:mullewa}
\end{figure*}

In this experiment, we ran all unsupervised methods over a real-life hyperspectral scene. Since there is no ground-truth segmentation of the Mullewa dataset, we qualitatively compare the selected methods in Fig.~\ref{fig:mullewa}. Here, we present the segmentations obtained using all unsupervised techniques over (i)~full HSI, and (ii)~reduced HSI (this reduction was performed with the approach which was the best over all benchmarks for the corresponding segmentation algorithm). We can appreciate that $k$-means and 3D-CAE give much more detailed segmentation (see example regions annotated with the white and black arrows in the GM visualization). It indicates that those regions are ``heterogeneous'' and manifest subtle spectral variations. This observation can trigger more detailed in-situ measurements (performed in precise locations), hence allow us to better understand the scanned regions and their critical characteristics. As previously, the execution time of 3D-CAE was much longer than other methods (Table~\ref{tab:ranking})---this issue can be tackled by more aggressive pre-processing (e.g.,~band selection), parallel GPU training or by applying early stopping conditions to both training phases of 3D-CAE.

\section{Conclusion}\label{sec:conclusions}

\begin{figure}[ht!]
\centering
\newcommand{\mywidth}{0.33}
\hspace*{-0.1cm}
\setlength\tabcolsep{1pt}
\begin{tabular}{cccc}
  \includegraphics[width=0.94\columnwidth]{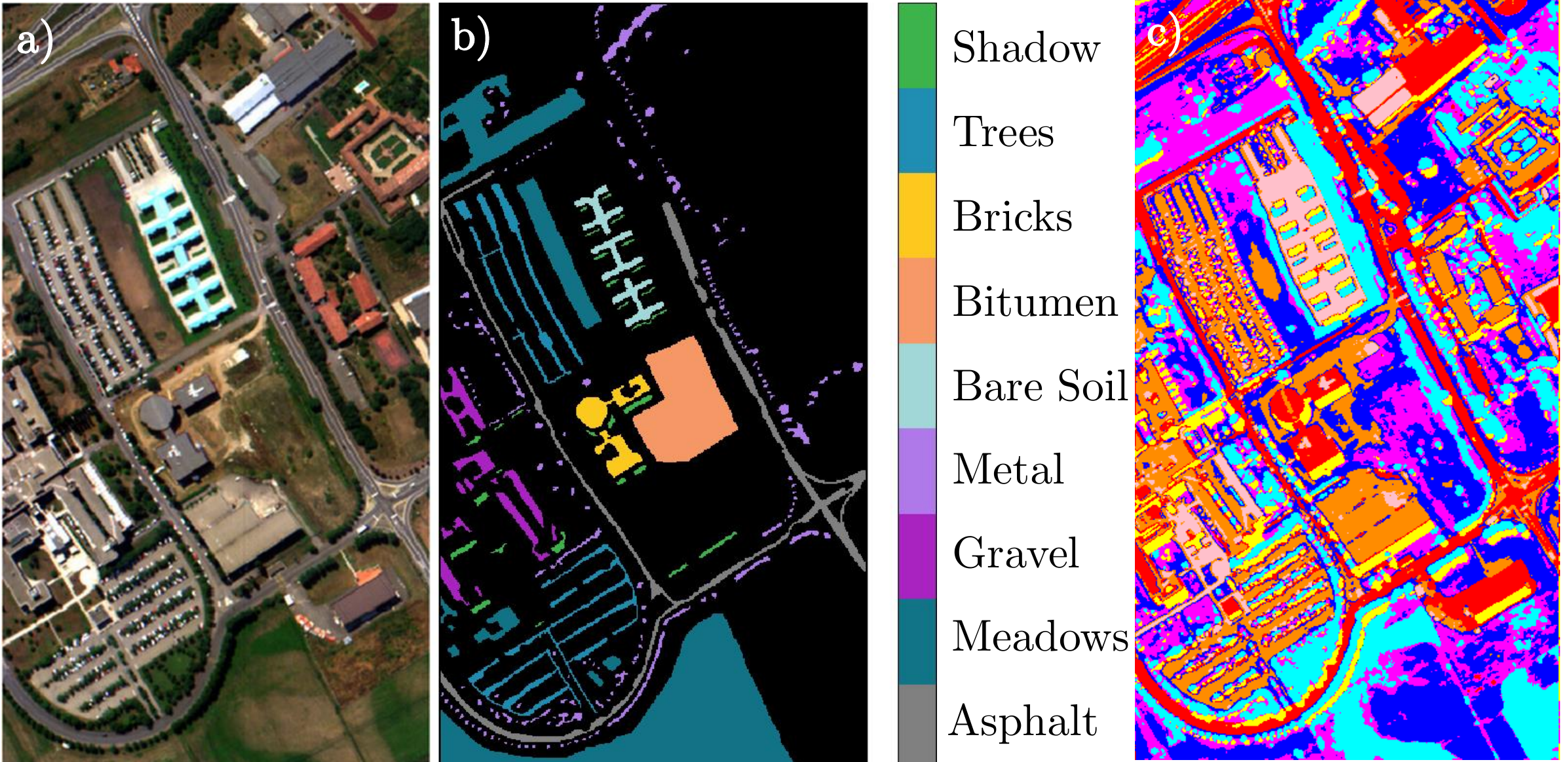}\\
\end{tabular}
	\caption{Unsupervised segmentation offers new possibilities of unrevealing information captured within newly acquired HSI and existent benchmarks. This example shows a)~the PU false-color scene, its b)~ground truth (black color is ``unknown class''), and c)~our full 3D-CAE segmentation which is not only very detailed, but also sheds new light on those ``unknown'' objects.}
	\label{fig:example_pu}
\end{figure}

We proposed a new deep learning-powered unsupervised HSI segmentation algorithm which exploits 3D convolutional autoencoders to learn embedded featues, and a clustering layer to segment an input image using the learned representation. Our experimental study, performed over benchmark and real-life HSI revealed that our approach delivers consistent and high-quality segmentation without any prior class labels. Such unsupervised techniques offer new possibilities to understand the acquired HSI---they can be used to: (i)~enable practitioners to generate ground-truth HSI data in affordable time even for very large scenes (unsupervised segmentation of an input HSI would be reviewed and fine-tuned if necessary), (ii)~perform anomaly detection within a captured region by analyzing unexpected heterogeneous parts of the segmentation map (e.g.,~a wheat farmland should be moderately homogeneous, and any deviation may be alarming), and to (iii)~see beyond the current ground-truth HSI (Fig.~\ref{fig:example_pu}). Although our method is computationally expensive, its execution time can be greatly decreased by the initial HSI reduction, applying early stopping conditions in both training phases, performing the parallel training (using multiple GPUs) and optimizing the hyper-parameters of the deep network architecture (e.g.,~decreasing the number of kernels)---it constitutes our current work.

\ifCLASSOPTIONcaptionsoff
  \newpage
\fi

\bibliographystyle{ieeetran}
\bibliography{IEEEabrv,ref_all}

\end{document}